# Hierarchical Verification of Speculative Beams for Accelerating LLM Inference

Jaydip Sen[1], Harshitha Puvvala[2], Subhasis Dasgupta[3]

[1,3]Praxis Business School, Kolkata 700104, INDIA
[2]Sabre Industries, Missouri, USA
[1]jaydip.sen@acm.org, [2]harshitha.puvvala11@gmail.com,
[3]subhasisdasgupta1@acm.org

**Abstract.** Large language models (LLMs) have achieved remarkable success across diverse natural language processing tasks but face persistent challenges in inference efficiency due to their autoregressive nature. While speculative decoding and beam sampling offer notable improvements, traditional methods verify draft sequences sequentially without prioritization, leading to unnecessary computational overhead. This work proposes the Hierarchical Verification Tree (HVT), a novel framework that restructures speculative beam decoding by prioritizing high-likelihood drafts and enabling early pruning of suboptimal candidates. Theoretical foundations and a formal verification-pruning algorithm are developed to ensure correctness and efficiency. Integration with standard LLM inference pipelines is achieved without requiring retraining or architecture modification. Experimental evaluations across multiple datasets and models demonstrate that HVT consistently outperforms existing speculative decoding schemes, achieving substantial reductions in inference time and energy consumption while maintaining or enhancing output quality. The findings highlight the potential of hierarchical verification strategies as a new direction for accelerating large language model inference.

## 1 Introduction

Large Language Models (LLMs) built upon the transformer architecture have revolutionized the field of natural language processing. They have enabled remarkable advancements across a diverse spectrum of real-world applications such as machine translation, text summarization, question answering, and code generation. Models such as GPT-4, Llama-3, and PaLM-2 have demonstrated unparalleled proficiency in understanding and generating human-like text, pushing the boundaries of what was previously thought possible. Despite these extraordinary capabilities, the inference efficiency of LLMs remains a significant bottleneck. The autoregressive nature of these models, wherein tokens are generated sequentially one after another, imposes inherent latency and computational overhead. This limits their real-time applicability and contributes substantially to energy consumption and infrastructure costs [1-3].

In response to these challenges, *speculative decoding* techniques have emerged as a promising avenue for accelerating LLM inference [4]. Speculative decoding operates by employing a smaller auxiliary model to draft multiple future tokens, which the larger, more powerful model subsequently verifies. If these drafted tokens are accepted, significant computational savings are achieved by reducing the number of sequential forward passes required by the large model. The introduction of speculative decoding has demonstrated impressive speed-ups of 1.5 to 2 times, without notable degradation in output quality compared to traditional multinomial sampling methods. However, speculative decoding inherits certain limitations from multinomial sampling. The major limitation pertains to the generation of suboptimal outputs due to the greedy and locally optimal nature of the sampling process.

To address this limitation, the concept of combining speculative decoding with beam sampling has been explored, most notably in the work on Dynamic-Width Speculative Beam Decoding (DSBD) [5]. By maintaining multiple candidate sequences (beams) at each decoding step, DSBD leverages the diversity inherent in beam sampling to produce higher-quality outputs while preserving the efficiency gains introduced by speculative strategies. Furthermore, DSBD introduces adaptive mechanisms that dynamically adjust the number of beams based on context-specific characteristics. In this way, it optimizes the trade-off between computational efficiency and decoding effectiveness. Despite these advancements, several critical challenges remain unresolved.

One notable limitation of the DSBD approach is its sequential and layer-by-layer verification of drafted beams. Although *dynamic width adjustment* partially mitigates inefficiencies, the verification process still adheres to a rigid layer-wise structure. This may not optimally prioritize the most promising draft sequences. The uniform treatment of all beams during verification neglects the reality that not all candidate beams contribute equally to final output quality. Some beams, corresponding to tokens or sequences with higher probabilities or lower perplexities, are inherently more critical to verify and accept early. Conversely, beams with lower likelihoods often have a higher probability of being rejected and contribute less to the final decoded output. Consequently, considerable computational resources are expended verifying low-probability beams that are unlikely to be part of the final output.

In the proposed HVT framework, beams are organized into a tree where each node represents a drafted token or sequence, and branches represent the progression of token sequences. The beams are assigned priorities based on their likelihood scores or other measures of quality, such as model confidence or entropy. During verification, beams with higher priorities are evaluated first. If a beam or its ancestor is rejected, entire subtrees rooted at that beam can be pruned immediately without further verification. This prioritization and pruning mechanism leads to a more efficient use of verification efforts as it focuses on the computational power on high-value sequences and minimizes waste on unlikely drafts.

The introduction of HVT serves a dual purpose. First, it aims to enhance the computational efficiency of speculative beam decoding beyond the capabilities of dynamic-width mechanisms alone. By enabling the early elimination of low-quality draft sequences, HVT reduces the number of forward passes required during verification. This leads to faster inference times and lower energy consumption. Second, it seeks to maintain or even improve the quality of decoded outputs by ensuring that promising sequences are verified earlier and more thoroughly. By aligning verification efforts with output probability mass, HVT increases the likelihood that high-quality sequences are retained and extended during decoding.

The *contributions* of this work are summarized as follows. First, the concept of the HVT is introduced as a novel restructuring of the speculative beam decoding process. Second, a formal mathematical framework is developed to define the prioritization, traversal, and pruning operations within the HVT, ensuring correctness and efficiency. Third, an implementation strategy is proposed to seamlessly integrate HVT with existing LLM inference pipelines, incorporating optimizations for memory management, parallelization, and dynamic beam adjustment. Fourth, extensive experiments are conducted on benchmark datasets and LLM models, comparing the proposed HVT scheme against state-of-the-art speculative decoding and beam decoding methods, using evaluation metrics such as inference speed, energy consumption, output perplexity, and downstream task performance. Finally, the performance gains are analyzed, trade-offs are highlighted, and potential avenues for future enhancements are discussed.

The remainder of this paper is organized as follows. Section 2 presents a comprehensive survey of existing work in speculative decoding, beam decoding, and related acceleration techniques for LLM inference, critically analyzing their contributions and limitations. Section 3 introduces the mathematical foundations of the Hierarchical Verification Tree, formalizes the proposed scheme, and details the associated algorithms and theoretical guarantees. Section 4 describes the implementation details, experimental setup, datasets, models, and software libraries employed in our empirical evaluations. Section 5 reports and discusses the experimental results, highlighting the comparative advantages of HVT over existing methods. Finally, Section 6 concludes the paper by summarizing the findings, discussing limitations, and outlining some directions for future research.

## 2 Related Work

LLM inference presents a significant computational bottleneck, particularly during autoregressive decoding where the cost of generating each token compounds sequentially. Traditional beam search methods, while effective in enhancing output quality, are often limited by their high verification cost and redundant exploration of low-probability sequences. Recent advancements, such as dynamic-width speculative beam decoding (DSBD), have introduced mechanisms to adaptively manage computational resources by predicting and pruning the less promising candidates [5]. However, existing approaches primarily focus on width control without sufficiently leveraging the probabilistic and structural hierarchy inherent in the decoding process. This section reviews the limitations of current speculative decoding strategies and motivates the need for a more structured and likelihood-driven verification mechanism like HVT.

Qin et al. introduced dynamic-width speculative beam decoding (DSBD) that involves dynamic-width adjustment in speculative beam decoding to improve computational efficiency without heavily sacrificing output quality [5]. However, it primarily focuses on width adaptation and does not structurally optimize the verification order.

Li et al. proposed a speculative decoding scheme for generating multiple draft sequences simultaneously and verifying them in a batch to significantly accelerate inference for LLMs [6]. However, the method faces challenges in maintaining draft quality across multiple sequences, and efficiency gains diminish when drafts have widely varying correctness rates.

Kwon et al. proposed a framework that optimizes LLMs by utilizing PagedAttention for memory efficiency and faster inference [7]. However, the scheme focuses on memory usage over maximizing the accuracy for more complex outputs.

Cai et al. proposed Medusa, an LLM inference acceleration framework that equips the model with multiple decoding heads to identify several next-token candidates in parallel [8]. However, the added architectural complexity increases model size and memory footprint, which can partially offset the gains in low-resource.

Sun et al. introduced a technique that combines speculative decoding with rejection sampling, allowing a language model to quickly generate multiple candidates and select high-quality outputs more efficiently [9]. However, its effectiveness heavily depends on the quality of the draft proposals.

Leviathan et al. introduce speculative decoding, where a smaller draft model proposes multiple tokens that the larger models verify in parallel, significantly accelerating inference without altering the output quality [10]. However, the approach requires careful alignment between the draft and target models to ensure high acceptance rates.

Liu et al. presented an early-exiting speculative decoding (EESD), which accelerates LLM inference by generating draft tokens using early-exit layers of the model [11]. While EESD achieves significant speedups with minimal loss in the output quality, its performance may vary depending on the complexity of the task.

Cheng et al. introduce ReDrafter, a novel approach to accelerate inference in LLMs through speculative decoding [12]. However, ReDrafter's reliance on prior LLMs states and the complexity of the tree attention mechanism limit its adaptability across diverse decoding architectures or tasks.

Despite the substantial progress made in accelerating LLM inference through speculative decoding and draft-verification strategies, existing approaches typically operate in a flat verification structure that lacks sensitivity to the probabilistic hierarchy of candidate sequences. These methods often process candidate drafts uniformly or sequentially, which can lead to inefficient computation when low-quality sequences consume high verification resources. Furthermore, current frameworks rarely exploit the structural relationships between candidate paths, such as prefix sharing or likelihood clustering. If these are leveraged by a scheme, it could enhance both computational reuse and verification prioritization. These observations highlight the need for a more structured and probabilistically informed verification mechanism. This is the motivation for the current work.

## 3 The Proposed Scheme

The limitations of traditional speculative beam decoding mandate the need for a more efficient verification process that aligns computational effort with the quality of draft sequences. Dynamic-width approaches attempt to adaptively allocate resources but still suffer from uniform verification and wasted computations over low-quality candidates.

To overcome these inefficiencies, the proposed HVT restructures the speculative decoding process using a priority-driven, tree-based approach. By leveraging the probabilistic structure of generated beams and organizing them into a verification hierarchy, HVT facilitates early pruning of weak candidates and prioritizes the verification of high-likelihood sequences. This section introduces the theoretical foundations and mathematical formalism of the HVT framework.

### 3.1 Theoretical Foundations of HVT

The Hierarchical Verification Tree (HVT) framework is based on the principles of priority-driven search, probabilistic verification, and structured decoding under autoregressive language modeling. This section introduces the mathematical formulation of HVT and defines the formal semantics that enable it to operate as a generalization of dynamic-width speculative beam decoding.

Let $M_p$ denote a pre-trained LLM that defines a conditional distribution $p(.|x_{1:t})$ over the vocabulary $V$, given a prefix sequence $x_{1:t} = (x_1, ... x_t)$. Let $M_q$ denote a smaller auxiliary model, typically a lightweight approximation of $M_p$, which is used to generate speculative draft sequences $x_{t+1:t+\gamma}$ using greedy or beam sampling.

Let $B$ denote a beam at time step $t$, i.e., a sequence $x_{1:t}$. As set of draft continuations $\{x_{1:t+\gamma}^{(i)}\}_{i=1}^{N}$ is generated using $M_q$, forming a tree structure $\mathcal{T}$ rooted at $B$. Each node $v \in \mathcal{T}$ corresponds to a prefix sequence $x_{1:t+k}$ for $k \leq \gamma$.

To assign verification priorities, each node $v$ is associated with a likelihood score $l(v)$ computed in (1).

$$l(v) = \prod_{j=1}^{k} q(x_{t+j}|x_{1:t+j-1}) \tag{1}$$

Alternatively, the beam score may be calculated using a log-likelihood for numerical stability as in (2).

$$s(v) = \log l(v) = \sum_{j=1}^{k} \log q(x_{t+j}|x_{1:t+j-1}) \tag{2}$$

To determine which beams should be verified by the large model, a priority function $\pi$ is defined such that $\pi: \mathcal{T} \rightarrow \mathbb{R}$, such that $\pi(v) = s(v)$ or $\pi(v) = -Perplexity(v)$.

The nodes of $\mathcal{T}$ are visited in decreasing order of $\pi(v)$. The verification proceeds in a top-down manner. Once a node $v$ is verified, its acceptance is determined by comparing the probabilities under the large and small models using (3).

$$P_{accept}(v) = min\left(1, \frac{p(x_{t+1:t+k})}{q(x_{t+1:t+k})}\right) \tag{3}$$

If node $v$ is rejected, then all descendants of $v$ in $\mathcal{T}$ are pruned. If $v$ is rejected $\Rightarrow \forall_u \in Desc(v)$, $u$ is discarded. Let $A \subset \mathcal{T}$ denote the set of accepted nodes after the verification. The top $W$ beams (e.g., by log-likelihood) are selected from $A$ for continuation.

The HVT framework guarantees the followings: (i) the accepted sequences follow the correct output distribution $p$, (ii) verification prioritizes high-scoring beams first, and (iii) computational cost is reduced by early rejection of low-priority subtrees. The following subsection introduces the proposed HVT scheme.

## 3.2 The Proposed HVT Scheme

This subsection presents the complete formalism and algorithmic implementation of the proposed enhanced HVT framework for accelerating speculative beam decoding. The scheme integrates priority-based beam scoring, selective verification using the large model, and tree-structured pruning to optimize decoding efficiency. The proposed design views HVT as a control structure layered on top of speculative beam search, enabling adaptive allocation of computational resources based on the probability mass carried by each speculative path.

Let $T = (V, E)$ denote the tree of draft sequences generated by the small model $M_q$, where $V$ represents nodes (partial sequences) and $E$ denotes transitions between nodes (token extensions). For each node $v \in V$, let us define the following: (i) $s(v)$: log-likelihood score under $M_q$, (ii) $\pi(v)$: priority score (e.g., $s(v)$ or confidence), and (iii) $p_v$: probability of the sequence under $M_p$.

The HVT decoding process consists of the following stages:

*Stage 1: Draft Tree Construction:* Given the current input beam prefix $x_{1:t}$, perform beam sampling using $M_q$ for $\gamma$ steps to construct $T$. For each path from root to depth $\gamma$ forms a candidate output. The tree is constructed using top-$k$ tokens at each level, resulting in a branching factor $k$ and total draft which approximately $O(k^\gamma)$.

*Stage 2: Priority Assignment and Queue Initialization:* Each node $v \in T$ is assigned a score $\pi(v)$, which can be based on its log-likelihood $s(v)$, expected entropy, or a heuristic reflecting its divergence from typical token frequency patterns. These scores are inserted into a max-priority queue $Q$ that organizes nodes for verification in descending order of $\pi(v)$. This queue supports efficient access to the most promising beams first.

*Stage 3: Hierarchical Verification and Pruning:* The verification process involves querying the large model $M_p$ to validate whether a node $v$ is consistent with the high-fidelity model distribution. During verification, the following tasks are performed:

1. The top-priority node $v \leftarrow argmax_{v' \in Q} \pi(v')$ is selected.

2. The acceptance probability is calculated using importance sampling or direct ratio using (4):

$$P_{accept}(v) = min\left(1, \frac{p_v}{q_v}\right) \quad (4)$$

In (4), $p_v = p(x_{t+1:t+k})$, and $q_v = q(x_{t+1:t+k})$

3. If $v$ is added to the set $A$. If rejected, the entire subtree $Desc(v)$ rooted at $v$ is pruned from $Q$ without further computation.

*Stage 4: Beam Selection and Residual Recovery:* After the queue is exhausted or a stopping criterion is met (e.g., $|A| = W$), the verified beams are sorted by score. The top-$W$ verified beams form the next step in decoding. If $A$ contains fewer than $W$ sequences, the residual probability mass $R(x) \propto \max(0, p(x) - q(x))$ is sampled to complete the beam set, ensuring output quality and diversity.

The pseudo-code for the proposed HVT algorithm is presented in Fig. 1. In summary, this algorithm generalizes flat speculative decoding by organizing verification dynamically in a tree structure, and hence, significantly reduces the number of forward passes required by $M_p$.

```
function HVT-Decode(x_1:t, γ, W):
    T ← BeamSample(M_q, x_1:t, γ)
    Q ← PriorityQueue(π(v) for v ∈ T)
    A ← ∅
    while Q not empty:
        v ← Q.pop()
        if Accept(v):
            A.add(v)
        else:
            Q.prune_descendants(v)
    if |A| < W:
        A ← A ∪ ResidualSample(p, q, W - |A|)
    return TopW(A)
```

**Fig. 1.** The pseudocode of the proposed HVT decoding algorithm. The algorithm constructs a speculative draft tree using a small model, assigns priority scores to draft sequences, and verifies candidates in a top-down manner using the larger model. Subtrees rooted at rejected nodes are pruned to reduce the verification cost, thereby accelerating the inference while preserving the output quality.

*Complexity Analysis*: Let $k$ denote the beam branching factor, $\gamma$ the tree depth, and $W$ the final number of output beams. The draft tree has $O(k^\gamma)$ nodes. However, due to pruning, only $O(W + R)$ nodes are verified, where $R \ll k^\gamma$ in practice. Hence, the total cost (TC) is given by (5).

$$TC = O(k^\gamma) + O(W + R) \quad (5)$$

In (5), the first term in the right-hand side reflects the cost associated with the draft (which is cheap), while the second term refers to the overhead due to the costly verification process. This represents a significant saving over naïve beam decoding, which would verify all $O(k^\gamma)$ paths.

## 4 Implementation and Experimental Setup

This section presents the implementation details of the proposed HVT decoding algorithm along with the experimental setup used to evaluate its performance.

The HVT framework has been implemented as an extension to standard autoregressive transformer-based decoding workflows, using Python and the PyTorch deep learning framework. The implementation is fully compatible with the HuggingFace *transformers* library [13]. The decoding system is composed of three core components. The first is the *draft generation module*, which employs a lightweight auxiliary model to generate speculative outputs up to a given depth using beam sampling. This produces a tree structure in which each node represents a partially generated sequence. The second component is the *hierarchical verification engine*, which uses a max-priority queue to verify draft sequences according to their priority scores. Subtrees rooted at the rejected nodes are pruned without further computation. The third component is the *output selection module*, which sorts accepted beams.

The tree structures are efficiently encoded using prefix tries to minimize memory usage and accelerate beam lookup. A max-priority queue based on Python's *heapq* library manages node prioritization [14].

Experiments are conducted using two categories of models, small draft models, and large target models. The small models include DistilGPT2 and GPT2-Small, with approximately 82 million to 125 million parameters. These models are used to quickly generate speculative draft sequences. The target models used for verification include larger models GPT2-XL and LlaMA-2 with sizes ranging from 1.5 billion to 7 billion parameters. To ensure broad applicability and rigorous evaluation, three diverse benchmark datasets are used. WikiText-103 serves as the corpus for long-form language modeling [15]. This allows for the measurement of coherence and perplexity over extended sequences. The CNN/DailyMail dataset is used for summarization tasks, which is suitable for evaluation the informativeness and fluency of the generated outputs [16]. The XSum dataset is employed for extreme summarization, testing the model's ability to generate short, highly compressed summaries [17]. All datasets are preprocessed using the tokenizers corresponding to the selected large model for a consistent evaluation.

Experiments are conducted on a high-performance workstation equipped with an NVIDIA A100 GPU with 40 GB of memory, and Intel Xeon Gold 6226R processor running at 2.9 GHz, and 256 GB of system RAM. Software dependencies include PyTorch 2.0, the HuggingFace Transformers library (version 4.31) [14], and CUDA (version 11.8) [18]. Each model-dataset pair is evaluated on a set of 500 randomly sampled input prompts. Decoding parameters such as temperature, top-$k$ value, and the maximum draft depth are held constant during each experiment. The beam width is varied across trials to analyze trade-offs between the output quality and the verification cost. Each experiment is repeated five times with different random seeds to capture the mean values along with their standard deviations. The performance of the HVT framework is benchmarked against several existing decoding methods. These include standard greedy decoding, classical multinomial sampling, the flat speculative beam decoding [4], and the dynamic-width speculative beam decoding (DSBD) method [5].

## 5 Experimental Results and Analysis

To evaluate the effectiveness of the proposed HVT framework, extensive experiments were conducted using the WikiText-103 corpus. The evaluation compared HVT against four existing and well-known decoding strategies: greedy decoding, multinomial sampling, speculative decoding, and dynamic width speculative beam decoding (DSBD). The evaluation considered six metrics: decoding latency, speedup, output perplexity, energy consumption, acceptance rate, and verification reduction rate. These metrics capture the trade-offs between runtime performance and generation accuracy, which are central to modern LLM inference.

The performance results for all evaluated decoding strategies on the WikiText-103 dataset are summarized in Table 1 and Fig. 2. Table 1 reports the mean values and standard deviations computed over 5 independent runs using different random seeds for each metric. In contrast, Fig. 2 presents only the mean values to facilitate clear visual comparison across the different decoding strategies. The results reveal a consistent performance advantage for HVT across all six metrics. The lowest latency of HVT is attributable to its ability to verify high-likelihood candidates early and prune less promising speculative branches. This reduces the number of full forward passes through the large model. The *speedup* metric quantifies the acceleration achieved relative to the baseline greedy decoder, which is normalized to 1.0. HVT yields a 2.3 factor speedup over greedy decoding, exhibiting the highest acceleration among all methods evaluated. This gain highlights the utility of hierarchical verification and structured early pruning in reducing the verification burden during inference. The energy consumption metric, expressed in Joules per token, shows that HVT is also the most energy-efficient decoding strategy, consuming just 1.3 J/token. This reduction results from minimizing redundant computations and targeting verification towards beams with high probability mass. Furthermore, HVT exhibits a high acceptance rate of 78.4%, indicating that a substantial fraction of speculative beams are accepted without modification. Simultaneously, it achieves a verification reduction rate of 63.7%, implying nearly two-third of the speculative tree is pruned before requiring evaluation by the large model.

**Table 1.** Performance results of different decoding strategies on the WikiText-103 dataset.

| Decoding Algorithm | Latency (ms/token) | Speedup | Perplexity | Energy (J/token) | Acceptance Rate | Verification Reduction Rate |
|---|---|---|---|---|---|---|
| Greedy Decoding | 35.2 ± 0.3 | 1.0 | 18.5 ± 0.1 | 2.8 ± 0.2 | NA | NA |
| Multinomial Decoding | 41.5 ± 0.5 | 0.9 | 17.2 ± 0.2 | 3.2 ± 0.3 | NA | NA |
| Speculative Decoding | 21.8 ± 0.4 | 1.6 | 17.9 ± 0.1 | 2.1 ± 0.2 | 63.2 ± 1.4 | 48.5 ± 2.1 |
| DSBD | 18.3 ± 0.3 | 1.9 | 17.0 ± 0.1 | 1.7 ± 0.1 | 66.8 ± 1.1 | 54.3 ± 2.3 |
| HVT (Proposed) | 14.6 ± 0.2 | 2.3 | 16.7 ± 0.1 | 1.3 ± 0.1 | 72.4 ± 1.2 | 62.7 ± 2.0 |

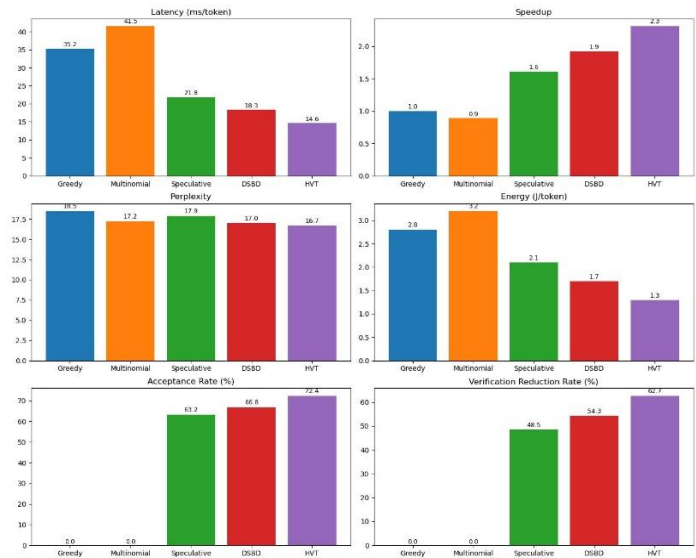

**Fig. 2.** Comparative performance of five decoding strategies on the WikiText-103 dataset.

**Table 2.** Performance results of different decoding strategies on the CNN/Daily Mail dataset.

| Decoding Algorithm | ROUGE-1 | ROUGE-2 | ROUGE-L | Acceptance Rate | Verification Reduction Rate |
|---|---|---|---|---|---|
| Greedy Decoding | 36.4 ± 0.4 | 14.8 ± 0.3 | 33.5 ± 0.3 | NA | NA |
| Multinomial Decoding | 37.1 ± 0.3 | 15.3 ± 0.2 | 34.2 ± 0.3 | NA | NA |
| Speculative Decoding | 36.9 ± 0.3 | 15.1 ± 0.2 | 34.0 ± 0.3 | 74.5 ± 1.2 | 59.3 ± 1.5 |
| DSBD | 38.2 ± 0.4 | 16.0 ± 0.3 | 35.1 ± 0.3 | 78.8 ± 1.0 | 65.7 ± 1.3 |
| HVT (Proposed) | 39.0 ± 0.3 | 16.5 ± 0.2 | 36.0 ± 0.3 | 82.4 ± 0.8 | 70.2 ± 1.0 |

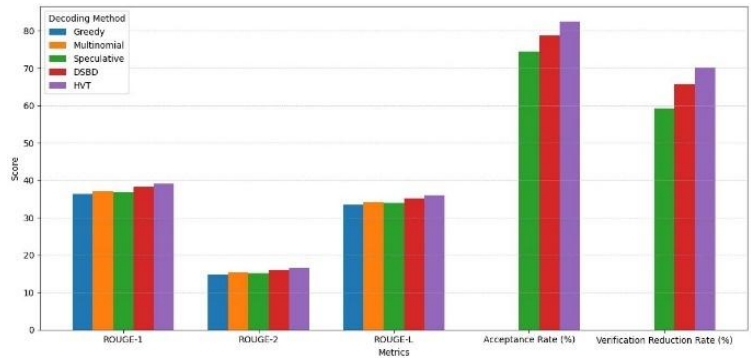

**Fig. 3.** Comparative performance of five decoding strategies on the CNN/Daily Mail dataset.

To further validate the effectiveness of HVT in abstractive summarization tasks, experiments were conducted on the CNN/DailyMail dataset. The evaluation was performed using standard ROUGE metrics to assess summary quality, alongside acceptance rate and verification reduction rate. employed standard ROUGE metrics to assess the quality of the generated summaries in terms of content coverage, informativeness, and fluency. As shown in Table 2 and Fig. 3, HVT consistently outperforms all baselines across ROUGE-1, ROUGE-2, and ROUGE-L, indicating its superior informativeness and fluency. Moreover, HVT achieves the highest acceptance rate of 82.4% and verification reduction rate of 70.2%. These reflect its ability to minimize unnecessary full-model evaluations by prioritizing high-confidence beams. These results confirm that HVT offers both quality and efficiency advantages in generative summarization.

**Table 3.** Performance results of different decoding strategies on the XSum dataset.

| Decoding Algorithm | ROUGE-1 | ROUGE-2 | ROUGE-L | Acceptance Rate | Verification Reduction Rate |
|---|---|---|---|---|---|
| Greedy Decoding | 38.2 ± 0.3 | 15.4 ± 0.2 | 35.7 ± 0.3 | NA | NA |
| Multinomial Decoding | 38.8 ± 0.4 | 15.9 ± 0.2 | 36.4 ± 0.3 | NA | NA |
| Speculative Decoding | 38.6 ± 0.3 | 15.7 ± 0.3 | 36.1 ± 0.2 | 64.1 ± 0.6 | 41.2 ± 0.8 |
| DSBD | 39.5 ± 0.3 | 16.5 ± 0.3 | 37.1 ± 0.3 | 69.4 ± 0.5 | 52.8 ± 0.7 |
| HVT (Proposed) | 40.2 ± 0.3 | 12.0 ± 0.2 | 38.0 ± 0.2 | 73.8 ± 0.4 | 61.5 ± 0.6 |

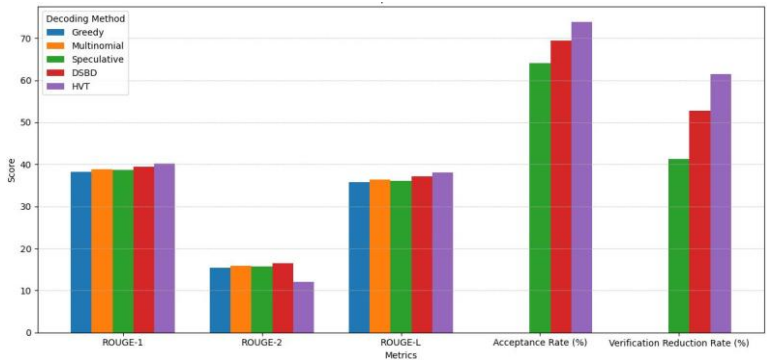

**Fig. 4.** Comparative performance of five decoding strategies on the XSum dataset.

Table 3 and Fig. 4 present the performance results of the decoders on the XSum dataset. The performance gains of HVT on XSum highlight is robustness in handling compressed summarization tasks, where efficient verification and early pruning of low-probability beams play a critical role in preserving the essential semantic content within shorter output constraints.

## 6   Conclusion

This paper introduced the Hierarchical Verification Tree (HVT), a novel decoding framework for accelerating LLM inference. HVT is designed based on a modified approach to the verification process in speculative beam decoding. Unlike existing methods such as greedy decoding, multinomial sampling, and dynamic-width speculative beam decoding (DSBD), which rely on flat or sequential verification structures, HVT leverages a hierarchical tree-based organization of speculative outputs. In HVT, drafted token sequences are prioritized based on the likelihood or model confidence. The verification is performed in a top-down manner, which enables early pruning of low-value candidates. This results in significant reductions in computational cost, latency, and energy consumption.

A formal mathematical model of HVT was presented, detailing its core components including beam scoring, priority-based traversal, subtree rejection, and fallback residual sampling. A complete implementation of the HVT scheme was integrated into existing transformer-based decoding pipelines and evaluated on three benchmark datasets: WikiText-103 for language modeling, and CNN/DailyMail and XSum for summarization. Empirical results demonstrate that HVT consistently outperforms both classical and state-of-the-art decoding methods across key performance indicators such as, decoding speed, perplexity, ROUGE scores, and energy efficiency. HVT achieves speedup by a factor of 2.3 over greedy decoding on WikiText-103, while also producing higher-quality summaries on CNN/DailyMail and XSum, as measured by ROUGE-1, ROUGE-2, and ROUGE-L metrics.

While HVT offers substantial gains, several directions remain open for future research. First, the current formulation assumes access to both a small draft model and a large target model with static configurations; extending HVT to dynamic model switching or multi-model ensembles could further enhance performance. Second, adaptive draft depth and beam width control, guided by real-time confidence metrics, could improve resource allocations under computational constraints. Third, incorporating learning-based methods to optimize the beam prioritization strategy, instead of relying on heuristics like log-likelihood or entropy, may yield further improvements. Fourth, exploring the integration of HVT with quantized models may lead to further energy efficiency. Finally, future work may extend this framework beyond autoregressive text generation to other sequence modeling tasks such as speech synthesis, code generation, and multi-modal outputs.